\documentclass[letterpaper,10pt,conference]{ieeeconf}  

\IEEEoverridecommandlockouts                              

\usepackage{mymacros}
\usepackage{times} 
\usepackage{graphicx}
\usepackage{caption}
\usepackage[T1]{fontenc}
\usepackage{color}
\usepackage{pifont}

\let\labelindent\relax
\usepackage{enumitem}

\title{\LARGE \bf
Object Segmentation from Open-Vocabulary Manipulation Instructions Based on Optimal Transport Polygon Matching with \\ Multimodal Foundation Models
}

\author{
    Takayuki Nishimura, Katsuyuki Kuyo, Motonari Kambara and Komei Sugiura
\thanks{
    \small The authors are with Keio University, 3-14-1 Hiyoshi, Kohoku, Yokohama, Kanagawa 223-8522, Japan.
    {\tt\small t-nishimura@keio.jp}
}
}

\begin{document}

\maketitle
\thispagestyle{empty}
\pagestyle{empty}

\begin{abstract}
We consider the task of generating segmentation masks for the target object from an object manipulation instruction, which allows users to give open vocabulary instructions to domestic service robots.
Conventional segmentation generation approaches often fail to account for objects outside the camera's field of view and cases in which the order of vertices differs but still represents the same polygon, which leads to erroneous mask generation.
In this study, we propose a novel method that generates segmentation masks from open vocabulary instructions.
We implement a novel loss function using optimal transport to prevent significant loss where the order of vertices differs but still represents the same polygon.
To evaluate our approach, we constructed a new dataset based on the REVERIE dataset and Matterport3D dataset. 
The results demonstrated the effectiveness of the proposed method compared with existing mask generation methods.
Remarkably, our best model achieved a +16.32\% improvement on the dataset compared with a representative polygon-based method.
\end{abstract}

\section{Introduction
}
In modern aging societies, the demand for assistance and support in daily life is increasing; however, there is a feared shortage of home caregivers.
As a possible solution, domestic service robots (DSRs) capable of providing physical assistance for caregiving are attracting significant attention.
Allowing care recipients to give instructions to DSRs in natural language could greatly increase convenience.
\begin{figure}[t]
    \vspace{2mm}
    \centering
    \includegraphics[width=\linewidth]{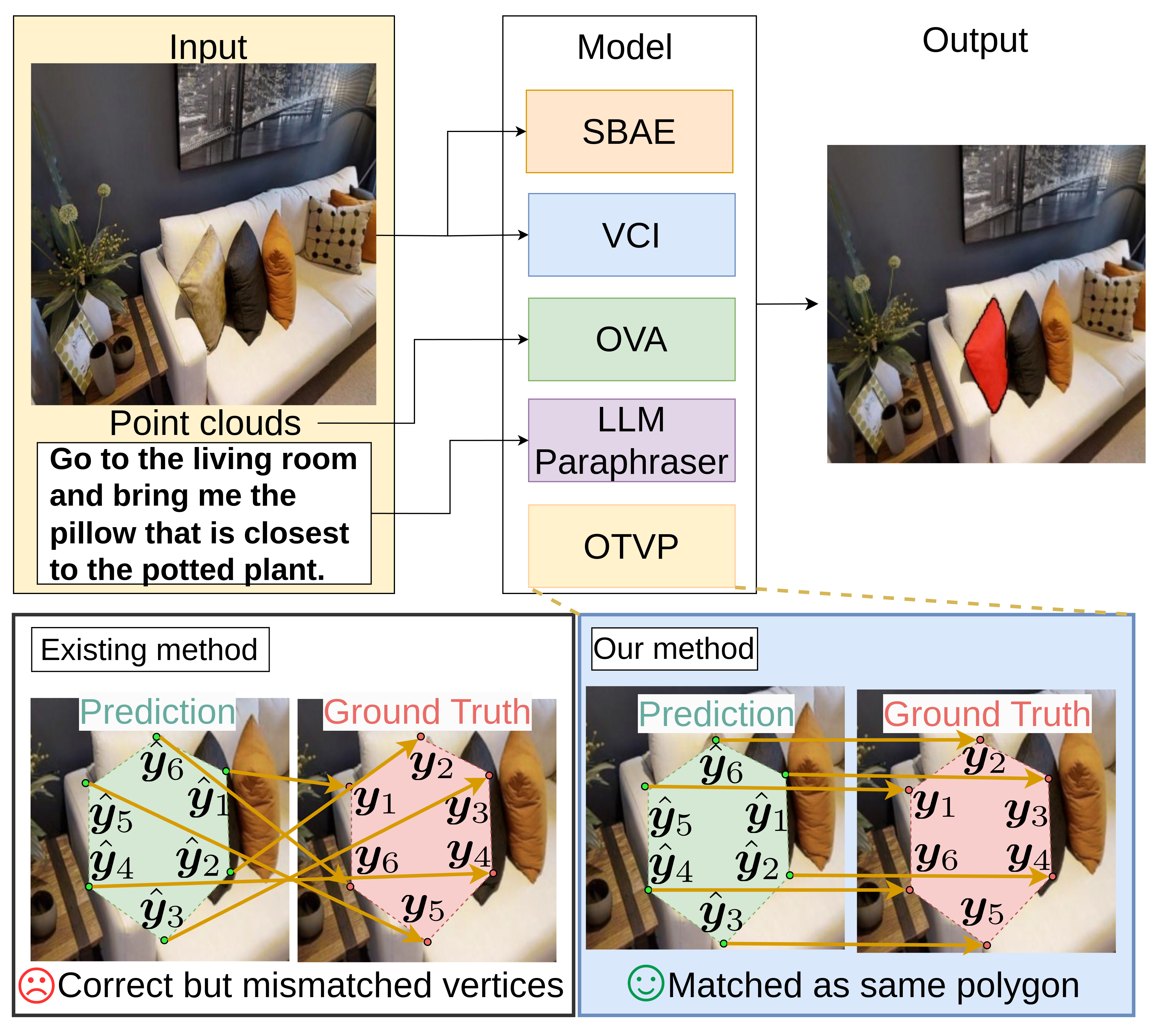}
    \caption{\small Overview of our method. Our method generates a polygon-based segmentation mask for the target object of a given instruction and image. We introduce the Polygon Matching Loss. The LLM Paraphraser, SBAE, OVA, VCI, and OTVP are explained in Section~\ref{methods}.}
    \label{fig:eye_catch}
    \vspace{-6mm}
\end{figure}
However, such instructions sometimes incorporate out-of-vocabulary words, complex referring expressions and redundant phrases.
This complexity makes it challenging for DSRs to understand such instructions and identify target objects.

In this study, we focus on the task of generating segmentation masks of the target object given open vocabulary instructions related to object manipulation.
This task is important because it is convenient for users if robots can understand and execute object manipulation based on natural language instructions.
For instance, given the instruction, ``Go to the living room and bring me the pillow that is closest to the potted plant,'' it is required to generate a segmentation mask for the pillow that is closest to the potted plant.
Segmentation masks are more desirable than bounding boxes for object manipulation because it is desirable to accurately predict the position and shape of target objects. 

Although our target task is closely related to the referring expression segmentation (RES) task\cite{hu2016segmentation}, the instructions in our target task often involve two or more sentences.
Therefore, it is necessary to identify the object by considering complex relationships between vision and language.
Thus, our target task is more challenging than the simple RES task.
For instance, consider the instruction ``Go downstairs to the open living room with the white fireplace and straighten out the book display next to it.'' 
Referring only to ``the book display next to it'' is too ambiguous to appropriately identify the target object.
In this case, ``the white fireplace'' indirectly modifies the target object and is important for understanding the instruction.

Although many models \cite{yang2022lavt,wang2021cris,iioka2023mdsm} have been successfully applied to the RES task, most of them do not fully handle multiple sentences.
Furthermore, they are also unable to handle the referring expressions of objects that exist outside the camera's view.
Recently, some studies show that proposed polygon-based mask generation methods can achieve shorter inference times compared with traditional pixel-based mask generation methods\cite{peng2020deep,liu2021dance,liu2023polyformer,wang2021cris}.
However, most of them cannot account for cases in which the order of vertices differs while still representing the same polygon.

In this study, we propose a model that generates a segmentation mask for the target object specified in a natural language instruction.
One of the main differences between our method and existing methods is the introduction of the Polygon Matching Loss (PML), which uses optimal transport for vertex matching.
Another significant difference is the introduction of Open-Vocabulary 3D Aggregator (OVA), which handles open-vocabulary multimodal features for objects that exist outside the camera’s field of view.

Introducing PML enables the model to handle cases in which the order of vertices differs while still representing the same polygon.
As a result, we train the model to predict the appropriate masks regardless of the vertex order, thereby enabling effective training.
Additionally, we expect the OVA to enhance the association of open-vocabulary multimodal features with referring expressions that refer to objects that exist outside the camera's field of view.

A summary of our key contributions is as follows:
\begin{itemize}
    \item To train the model efficiently, we introduce the Optimal Transport Vertex Predictor (OTVP), with the PML, which uses optimal transport for vertex matching.
    \item We introduce the OVA to obtain open-vocabulary multimodal features for objects that exist outside the camera's field of view.
    \item We introduce the Segment-Based Attentional Enhancer (SBAE), which uses segmentation images to enhance the understanding of object shapes and their spatial relationships.
\end{itemize}

\begin{figure*}[t]
    \centering
    \vspace{2mm}
    \includegraphics[width=0.95\linewidth]{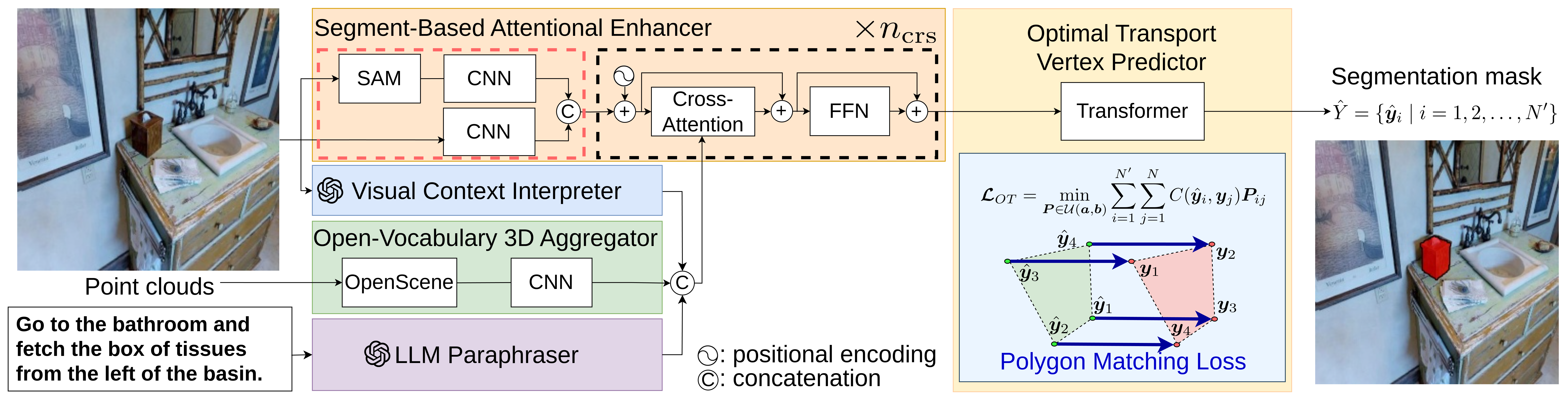}

    \caption{\small Proposed method framework. The proposed method consists of five main modules: LLM Paraphraser, SBAE, OVA, Visual Context Interpreter (VCI), and OTVP.
    $C\left(\cdot, \cdot\right)$, SAM, OpenScene represent the cost function, Segment Anything Model\cite{Kirillov_2023_ICCV}, and Open Scene\cite{Peng2023OpenScene}, respectively.}
    \label{fig:model}
    \vspace{-8mm}
\end{figure*}
\vspace{-2mm}
\section{
    Related Work
}
In the field of multimodal language processing, numerous studies have been conducted\cite{UPPAL2022149, chen2023vlp, gu2022vision, zhu2023survey} and multimodal large language models (LLMs) have led to notable successes\cite{gpt4v,dai2023instructblip}.
%
Multimodal LLMs have progressed rapidly, and have been successfully applied to task planning \cite{Driess2023palme,brohan2023can} and the generation of code for action sequences \cite{vemprala2023chatgpt,liang2023code}.
These approaches have been widely applied in robotics \cite{miyazawa2023survey, xiao2023robot, kawaharazuka2024real, zeng2023large}.

Referring expression comprehension (REC) and RES are two major tasks in which models are required to predict specific regions within images based on referring expressions (e.g., \cite{Yu_2018_CVPR,Luo_2020_CVPR,Ye_2019_CVPR}).
REC often requires predicting the rectangular regions of target objects given images and referring expressions\cite{kamath2021mdetr,deng2021transvg,UNINEXT}. 
Therefore, in this study, we focus on segmentation generation tasks rather than REC tasks.

Most RES models predict the pixel-level masks of target objects\cite{yang2022lavt,wang2021cris,huang2020referring,zou2023segment,GRES}, whereas some predict the vertices of a polygon that represents the target object\cite{zhu2022seqtr,liu2023polyformer}.
Although these polygon-based mask generation methods are similar to our method, they cannot account for cases in which the order of vertices differs but represents the same polygon.
Unlike them, we introduce polygon matching with optimal transport in mask generation to achieve efficient training.
As a result, despite appropriately predicting the set of vertices, most existing methods do not consider polygons that are similar, which results in the inefficient training of the model.

Additionally, several studies have been conducted aimed at referring expression understanding for DSRs \cite{iioka2023mdsm,kaneda2024learning,korekata2023switch,karamcheti2023voltron}. 
These tasks involve decomposing high-level instructions into atomic actions and executing them \cite{karamcheti2023voltron,lynch2023interactive,chen2023polarnet}, and identifying the target objects specified in the instruction sentences \cite{kaneda2024learning,iioka2023mdsm,homerobot,parashar2023slap}.
The authors of \cite{iioka2023mdsm} proposed MDSM, which is a two-stage segmentation model designed to refine masks generated by DDPM \cite{ho2020ddpm}.
Unlike MDSM, our method handles information about objects that exist outside the camera's field of view.


Many datasets for referring expression understanding have been proposed RefCOCO\cite{kazemzadeh2014referitgame}, RefCOCO+\cite{yu2016modeling}, G-Ref\cite{mao2016generation}.
In the field of robotics, datasets that contain natural language instruction sentences \cite{hatori2018interactively,qi2020reverie,shridhar2020alfred} are used for multimodal language understanding tasks. 
These datasets focus on object manipulation tasks within an indoor environment.
\cite{hatori2018interactively,qi2020reverie} are notable studies because they were based on real-world data.
In particular, the instruction sentences included in the REVERIE dataset\cite{qi2020reverie} often consist of multiple sentences, which make the task of identifying the target object particularly challenging.
\vspace{-1mm}
\section{Problem Statement
}
\vspace{-1mm}
In this study, we focus on the task that involves generating the segmentation mask of the target object from an image of the indoor environment, 3D point clouds, and an instruction related to object manipulation.
We define this task as the Object Segmentation from Manipulation Instructions-3D (OSMI-3D) task.
In this task, the model should generate a segmentation mask for the target object indicated in the instruction.
Fig.~\ref{fig:eye_catch} shows a typical input of the OSMI-3D task.
The goal is to generate a mask, which is indicated by the red area, given an instruction such as ``Go to the living room and bring me the pillow that is closest to the potted plant.''

We define the inputs and an output as follows:
\begin{itemize}
    \item \textbf{Inputs:} an image, 3D point cloud, and an instruction sentences.
    \item \textbf{Output:} a pixel-wise segmentation mask of the target object indicated in the instrucion.
\end{itemize}
In this study, we do not assume cases in which there are multiple target objects or no target object in a single image.
We use mean intersection over union (mIoU) and precision as the evaluation metrics.
\section{Proposed Method
\label{methods}
}
\vspace{-1mm}

The proposed method predicts a segmentation mask for the target object referred to in the given object manipulation instructions.
Our key contributions are as follows:
\begin{itemize}
    \item To train the model efficiently, we introduce the Optimal Transport Vertex Predictor (OTVP), with the PML, which uses optimal transport for vertex matching.
    \item We introduce the OVA to obtain the open-vocabulary multimodal features of objects that exist outside the camera's field of view.
    It can handle their correspondence with referring expressions.
    \item We introduce the SBAE to enhance the understanding of attributes, such as shape and spatial relationships, based on the segmentation images.
\end{itemize}
Fig.~\ref{fig:model} shows the overview of the proposed method. It consists of five main modules: LLM Paraphraser, SBAE, OVA, Visual Context Interpreter (VCI) and OTVP.
Our method, particularly the proposed PML, can be widely applied to polygon-based mask generation models\cite{liu2023polyformer,zhu2022seqtr}.

The inputs are defined as $\bm{x}=\{\bm{x}_{\mathrm{img}},X_{\mathrm{pcl}},\bm{x}_{\mathrm{inst}}\}$ where $\bm{x}_{\mathrm{img}} \in \mathbb{R}^{3 \times H \times W}$, $X_{\mathrm{pcl}} = \{ \bm{\xi}_i \mid i = 0, 1, 2, \ldots, N_{\mathrm{pcl}} \}$ and $\bm{x}_{\mathrm{inst}} \in \{0, 1\}^{v \times l}$ denote an image, 3D point clouds and an instruction sentences tokenized as a one-hot vector, respectively.
Note that $H$, $W$, $\bm{\xi}_i$, $N_{\mathrm{pcl}}$, $v$ and $l$ denote the height of the image, width of the image, $i$-th point, total number of points in a point cloud, vocabulary size and max token length of the instruction, respectively.
\vspace{-2mm}
\subsection{LLM Paraphraser}
Unlike the RES task, the OSMI-3D task often involves two or more sentences. 
As shown later, typical RES models do not handle such cases and may focus on phrases unrelated to the target object. 
To improve the understanding of phrases associated with the target object, we introduce the LLM Paraphraser.
Specifically, the LLM Paraphraser combines several sentences  and summarizes referring expressions related to the target object into a single sentence.

LLM Paraphraser takes $\bm{x}_{\mathrm{inst}}$ as input.
It summarizes $\bm{x}_{\mathrm{inst}}$ using an LLM (GPT-3.5-turbo\cite{gpt35turbo}).
For example, when $\bm{x}_{\mathrm{inst}}$ is ``Go to the dining table. Then pick up the candle on the right, '' the sentence ``Pick up the right candle on the dining table.'' is obtained. 
We embed the sentence into the language features $\bm{h}_{\mathrm{llp}} \in \mathbb{R}^{d_{\mathrm{llp}}}$ using the text-embedding-ada-002\cite{adatxtembeddingada002}, where $d_{\mathrm{llp}}$ is the number of feature dimensions.
The output of LLM Paraphraser is $\bm{h}_{\mathrm{llp}}$.
\vspace{-2mm}
\subsection{Visual Context Interpreter}
Previous RES studies can be mainly divided into two approaches for extracting image features: using image encoders (e.g.,  ResNet\cite{He2015DeepRLresnet} and ViT\cite{dosovitskiy2020image}) to extract visual features such as texture and edges; and using multimodal image encoders (e.g., CLIP\cite{radford2021learning}, UNITER \cite{chen2020uniter}, and BLIP\cite{li2022blip}) to extract multimodal image features that are aligned with natural language. However, these features sometimes lack visual representations related to complex referring expressions (e.g., ``the second chair from the left in the first floor dining room that has the mirror hanging above the fireplace'') and spatial relationships (e.g., ``the hand towel on the towel rack to the left of the sink'').

To handle such complex visual representations, we introduce the VCI.
In the VCI, multimodal LLMs generate descriptions that include details such as the attributes of objects, their spatial relationships, and their complex relationships in referring expressions.
Furthermore, using multimodal LLMs, we can obtain additional common-sense knowledge that is not contained in the image alone.
For example, if the scene shows an open door and the outside is visible through the door, it is highly likely to be an entrance.

The inputs of VCI are $\bm{x}_{\mathrm{img}}$ and $\bm{x}_{\mathrm{inst}}$, and the output is the intermediate feature $\bm{h}_{\mathrm{vci}}\in \mathbb{R}^{d_{\mathrm{vci}}}$, where $d_{\mathrm{\mathrm{vci}}}$ represents the dimension.
First, we obtain a description of $\bm{x}_{\mathrm{img}}$ using a multimodal LLM (gpt-4-vision-preview \cite{gpt4v}).
We embed the description into $\bm{h}_{\mathrm{vci}}$ using the text-embedding-ada-002.
\vspace{-2mm}
\subsection{Open-Vocabulary 3D Aggregator}
Existing methods~\cite{iioka2023mdsm,yang2022lavt,zhu2022seqtr} often fail to identify the target object given the referring expressions of objects outside the field of view.
To address this issue, we introduce the OVA to enhance the understanding of referring expressions for objects outside the field of view.
This module aligns 3D point clouds with open-vocabulary multimodal features and links them to referring expressions.
As a result, it is expected to obtain information about objects outside the camera's field of view without the need to capture images from various angles.

In this module, the input is $X_{\mathrm{pcl}}$ and the output is the intermediate feature $\bm{h}_{\mathrm{ova}} \in \mathbb{R}^{d_{\mathrm{ova}}}$, where $d_{\mathrm{ova}}$ denotes the dimensionality.
First, from $X_{\mathrm{pcl}}$, we extract the $N_{\mathrm{near}}$ points closest to the position where $\bm{x}_{\mathrm{img}}$ was captured.
This subset is denoted by $X_{\mathrm{near}}$.
We use only $N_{\mathrm{near}}$ points because referring expressions often refer to objects around the target object, and using points at remote locations is not efficient.
We obtain multimodal features $\bm{h}'_{\mathrm{ova}} = f\left( X_{\mathrm{near}} \right)$.
Note that, $f\left( \cdot \right)$ denotes the application of pre-trained OpenScene\cite{Peng2023OpenScene}.
OpenScene embeds multimodal features into each point of a 3D point cloud using CLIP\cite{radford2021learning}.
Finally, we obtain the feature $\bm{h}_{\mathrm{ova}} = \mathrm{MaxPool}\left(\mathrm{Upsample}\left(\bm{h}'_{\mathrm{ova}}\right)\right)$ where $\mathrm{MaxPool}\left( \cdot \right)$ and $\mathrm{Upsample}\left( \cdot \right)$ denotes upsampling process and max pooling, respectively.

\subsection{Segment-Based Attentional Enhancer}
\begin{figure}[t]
    \centering
    \vspace{2mm}
    \includegraphics[width=\linewidth]{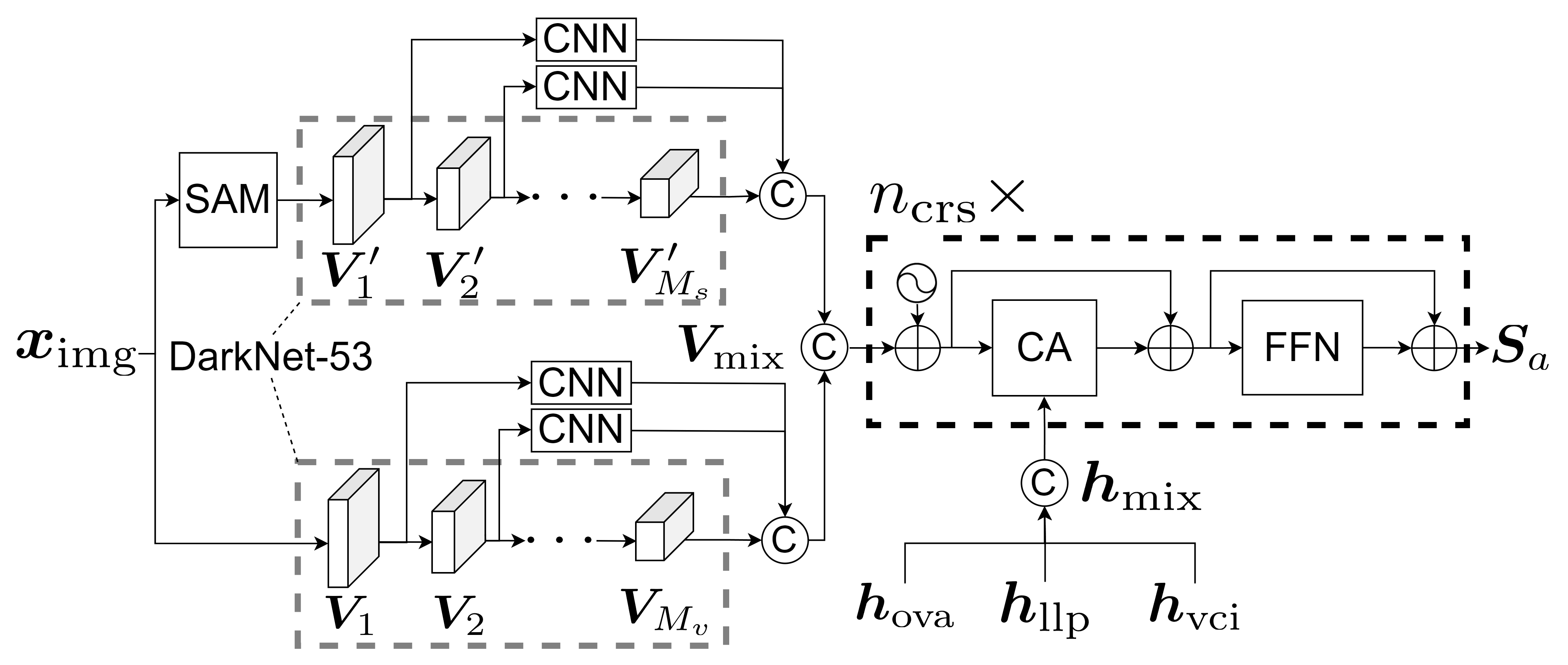}
    \caption{\small Structure of the SBAE. This enhances the understanding of object segment information, and fuses visual and linguistic features. CA, SAM and FFN represent cross-attention, the Segment Anything Model\cite{Kirillov_2023_ICCV} and a feed-forward network, respectively.}
    \label{fig:model_SBAE}
    \vspace{-7mm}
\end{figure}
Existing RES and OSMI models sometimes inappropriately predict the contours of objects. 
To address this, we introduce the SBAE to enhance the understanding of segment information related to target objects. 

Fig.~\ref{fig:model_SBAE} shows the structure of the SBAE.
It extracts image features at multiple resolutions from the $\bm{x}_{\mathrm{img}}$ and a segmentation image, and then integrates these features with $\bm{h}_{\mathrm{llp}}$, $\bm{h}_{\mathrm {vci}}$, and $\bm{h}_{\mathrm{ova}}$ from each module.
The inputs to this module are $\bm{x}_{\mathrm{img}}$, $\bm{h}_{\mathrm{llp}}$, $\bm{h}_{\mathrm {vci}}$ and $\bm{h}_{\mathrm{ova}}$.
First, we use the pre-trained SAM\cite{Kirillov_2023_ICCV} to obtain a segmentation image $\bm{s} \in \mathbb{R}^{3 \times H \times W}$ given $\bm{x}_{\mathrm{img}}$.
As shown in the figure, we obtain image features $\{\bm{V}_k \in \mathbb{R}^{H_k \times W_k \times C_k}\}_{k=1}^{M_v}$ from intermediate layers at $M_v$ different resolutions using DarkNet-53\cite{Wang_2021_CVPR} pre-trained on the MS-COCO\cite{lin2014microsoft} dataset.

Similarly, we obtain image features $\left\{\bm{V}'_l \in \mathbb{R}^{H_l \times W_l \times C_l} \right\}^{M_s}_{l=1}$ from $M_s$ types of intermediate layers from $\bm{s}$.
Note that $H$ and $W$ denote the height and width of the image feature maps of $\bm{V}_k$ and $\bm{V}'_l$, respectively, and $C_i$ denotes the number of channels of $\bm{V}_k$ and $\bm{V}'_l$.

As shown in Fig.~\ref{fig:model_SBAE}, we downsample each obtained $\bm{V}_k$ and $\bm{V}'_l$, and then obtain $\bm{V}_{\mathrm{mix}}$ by concatenating them in the channel dimension.
Furthermore, we obtain $\bm{h}_\mathrm{mix}$ by concatenating $\bm{h}_{\mathrm{llp}}$, $\bm{h}_{\mathrm {vci}}$ and $\bm{h}_\mathrm{ova}$ in the channel dimension, and $\bm{h}_\mathrm{mix}$ is downsampled.
Finally, we compute the cross-attention between $\bm{V}_{\mathrm{mix}}$ and $\bm{h}_\mathrm{mix}$ to obtain the intermediate feature $\bm{S}_{a} = f_{a}(\bm{V}_{\mathrm {mix}}, \bm{h}_{\mathrm {mix}})$.
Note that $f_a\left(\cdot, \cdot\right)$ denotes the cross-attention function.
Furthermore, $f_a\left(\cdot, \cdot\right)$ for any matrices $\bm{X}_{A}$ and $\bm{X}_{B}$ is defined as follows:
\begin{align*}
f_a(\bm{X}_{A}, \bm{X}_{B}) \hspace{-0.6mm}&=\hspace{-0.6mm} \mathrm{softmax}\left(\frac{(\bm{W}_{q}\bm{X}_{A})(\bm{W}_{k}\bm{X}_{B})^{\top}}{\sqrt{d}}\right)(\bm{W}_{v}\bm{X}_{B}),
\end{align*}
where $\bm{W}_q$, $\bm{W}_k$, and $\bm{W}_v$ are trainable weights, and $d$ is a scaling factor.
\vspace{-2mm}
\subsection{Optimal Transport Vertex Predictor}

The OTVP takes $\bm{S}_{a}$ as input. 
Output is $\hat{Y} = \{\hat{\bm{y}}_i \mid i = 1, 2, \ldots, N'\}$, where $\hat{\bm{y}}_i$ denotes the coordinate of a vertex.
The OTVP consists of a transformer encoder and transformer decoder.
The encoder and decoder consist of $n_{\mathrm{enc}}$ and $n_{\mathrm{dec}}$ transformer layers, respectively.
By inputting $\bm{S}_{a}$ into this encoder-decoder and applying linear projection, we obtain $\hat{Y}$.

Existing methods\cite{zhu2022seqtr,liu2023polyformer,liu2021dance,peng2020deep} often fail to account for cases in which the order of vertices differs but represents the same polygon.
The bottom-right diagram and bottom-left diagram in Fig.~\ref{fig:eye_catch} show an example.
The green and red shapes in the figure represent the predicted and correct masks, respectively. 
In the bottom-left diagram in Fig.~\ref{fig:eye_catch}, the two sets of vertices represent polygons of the same shape.
However, the order of vertices is different.
Despite appropriately predicting the set of vertices, most existing methods do not consider the polygons to be similar, which results in a significant loss.
This can lead to inefficient training of the model.

We introduce the PML $\mathcal{L}_{OT}$ to effectively address such cases using optimal transport. 
As shown in the bottom-right diagram in Fig.~\ref{fig:eye_catch}, it involves matching between $\hat{Y}$ and the set of vertices of reference mask $Y = \{\bm{y}_j \mid j = 1, 2, \ldots, N\}$ using optimal transport, where $N$ represents the number of vertices in the reference mask's set of vertices.

When $\hat{Y}$ and $Y$ are given, we can identify a transportation plan that transfers $\hat{Y}$ to $Y$ at the minimum transportation cost.
We regard $\hat{Y}$ and $Y$ as two discrete distributions 
$\alpha = \sum_{i=1}^{N'}\bm{a}_i\delta_{\hat{\bm{y}}_i}$ and $\beta = \sum_{j=1}^{N} \bm{b}_j\delta_{\bm{y}_j}$, respectively.
Note that $\delta_{\hat{\bm{y}}_i}$ and $\delta_{\bm{y}_j}$ represent the Dirac delta function centered on $\hat{\bm{y}}_i$ and $\bm{y}_j$, respectively.
The weight vectors $\bm{a}$ and $\bm{b}$ are normalized.
Using this, we define $\mathcal{L}_{OT}$ as follows:
\begin{align}
\label{eq:Lot}
  \mathcal{L}_{OT} = \min_{\bm{P} \in \mathcal{U}(\bm{a}, \bm{b})} \sum_{i=1}^{N'} \sum_{j=1}^{N} C(\hat{\bm{y}}_i, \bm{y}_j)P_{ij},
\end{align}
\begin{align*}
    \mathcal{U}(\bm{a}, \bm{b})\hspace{-1mm}=\hspace{-1mm}\left\{ \bm{P} \in \mathbb{R}^{{N'} \times {N}}\hspace{-1mm}\mid\hspace{-1mm} P_{ij} \geq 0, \bm{P} \mathbf{1}_{N'} = \bm{a}, \bm{P}^{\top} \mathbf{1}_{N} = \bm{b} \right\},
\end{align*}
where $\bm{1}_{N'}$ and $\bm{1}_{N}$ denote vectors of dimension $N'$ and $N$, respectively, with all components equal to 1.
Furthermore, $C(\bm{\hat{y}}_i, \bm{y}_j) = \|\bm{\hat{y}}_i - \bm{y}_j\|_2$ and ${P_{ij}}$ denote the transportation cost and transportation plan from $\bm{\hat{y}}_i$ to $\bm{y}_j$, respectively, where {$|| \cdot ||_2$} denotes the $L^2$ norm.
In this study, we compute \eqref{eq:Lot} efficiently with entropy regularization and subsequently apply the Sinkhorn algorithm\cite{cuturi2013sinkhorn}.

\section{
    Experimental Setup
    \label{exp_setup}
}


\subsection{
    Dataset
}
To the best of our knowledge, few datasets exist that contain all the information required for the OSMI-3D task.
Specifically, the OSMI-3D task requires a dataset that includes images of indoor environments, 3D point clouds, masks of target objects, and instruction sentences related to household tasks. 
The REVERIE dataset is a standard dataset for object localization collected from real indoor environments. Although it is closely related to our study, it does not include masks of the target objects, which makes it insufficient for the OSMI-3D task.
By contrast, the SHIMRIE dataset\cite{iioka2023mdsm} is another dataset for the OSMI task that includes masks of the target objects, but does not include 3D point clouds. Therefore, this dataset is also insufficient for our task.

From the above, instead of using these existing datasets, we constructed a new SHIMRIE-3D dataset based on the REVERIE \cite{qi2020reverie} and the Matterport3D \cite{chang2018matterport3d} datasets.
The SHIMRIE-3D dataset consists of images, 3D point clouds in Matterport3D, instruction sentences related to target objects, and polygon-based masks for those objects.
First, we collected instruction sentences from the REVERIE dataset. These instruction sentences were annotated by over 1,000 annotators using Amazon Mechanical Turk. 
The annotators were presented with animations of movement paths and randomly selected target objects. 
Then they were instructed to give an instruction related to object manipulation tasks on remote objects in real indoor environments.
The target object masks in the SHIMRIE-3D dataset were annotated semi-automatically with voxel-level class information about the objects and rectangular regions surrounding the target objects.
We used the voxel-level object class information contained in the Matterport3D dataset. We also used the rectangular area regions included in the REVERIE dataset.
We collected the 3D point clouds for the SHIMRIE-3D from the Matterport3D dataset.

The SHIMRIE-3D dataset included images with a resolution of $640 \times 480$.
The dataset contained 4,341 images, 11,371 instructions, and 11,371 corresponding masks for the target objects.
The dataset had a vocabulary size of 3,558, a total of 196,541 words, and an average sentence length of 18.8 words.
The dataset included a total of 11,371 samples.
The number of samples were 10,153, 856, and 362 for the training, validation, and test set, respectively.
We collected this dataset from 90 environments, which we split into seen and unseen environments according to the split defined in the REVERIE dataset.
The training set contained samples only from the seen environments.
The validation set contained 582 and 274 samples from the seen and unseen environments, respectively.
The test set contained samples only from the unseen environments.
We used the training and validation sets for updating parameters and selecting hyperparameters, respectively.
We used the test set to evaluate the performance of the model.
\vspace{-2mm}
\subsection{Parameter Settings}
For the cross-attention encoder in SBAE, we set the number of attention heads, number of layers, dimensionality of the feedforward network and input dimensionality as 8, $n_{\text{crs}}=2$, 1024 and 2048, respectively.
Similarly, for the transformer encoder and decoder in the OTVP, we set the number of attention heads, number of layers, dimensionality of the feedforward network, and input dimensionality as 8, 3, 1024 and 256, respectively.
We set $N=N'=N_{\text{near}}=10$, $M_v=M_s=3$.
We adopted the AdamW optimizer ($\beta_1$=0.9, $\beta_2$=0.999) for training with learning rate $5\times10^{-4}$, batch size 32, and dropout probability 0.1.
We used two loss functions: up to the 79th epoch, we used L1 loss between the predictions and the ground truth (GT), and from the 80th to the 90th epoch, we used $\mathcal{L}_{OT}$.


Our model had 320M trainable parameters and 960G total multiply-add operations.
We trained our model on a GeForce RTX 3090 with 24GB of memory and an Intel Core i9-12900K with 64GB of RAM.
The training time for the proposed method and inference time per sample were approximately 4 hours and 290 ms, respectively.
We computed mIoU on the validation set for each epoch. 
For the evaluation on the test set, we used the model with the maximum mIoU on the validation set.
\begin{figure*}[t]
    \centering
    \vspace{3mm}
    \captionsetup{}
    (\small i)
    \hspace{-0.5mm}
        \begin{minipage}{0.155\hsize}
            \centering
            \includegraphics[width=\linewidth]{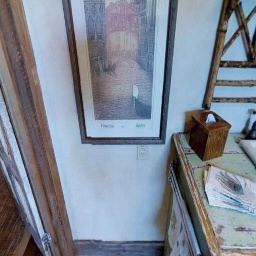}
        \end{minipage}
        \begin{minipage}{0.155\hsize}
            \centering
            \includegraphics[width=\linewidth]{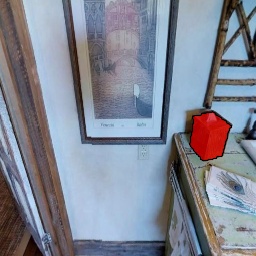}
        \end{minipage}
        \begin{minipage}{0.155\hsize}
            \centering
            \includegraphics[width=\linewidth]{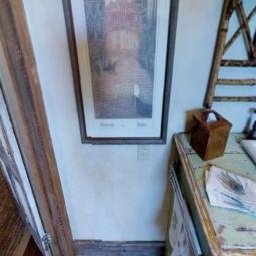}
        \end{minipage}
        \begin{minipage}{0.155\hsize}
            \centering
            \includegraphics[width=\linewidth]{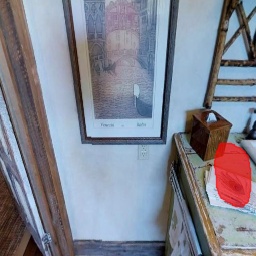}
        \end{minipage}
        \begin{minipage}{0.155\hsize}
            \centering
            \includegraphics[width=\linewidth]{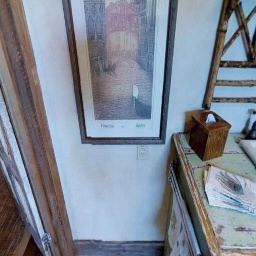}
        \end{minipage} 
        \begin{minipage}{0.155\hsize}
            \centering
            \includegraphics[width=\linewidth]{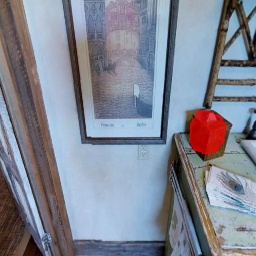}
        \end{minipage}\vspace{3mm}
        \vspace{3mm}
        (\small ii)
        \begin{minipage}{0.155\hsize}
            \centering
            \includegraphics[width=\linewidth]{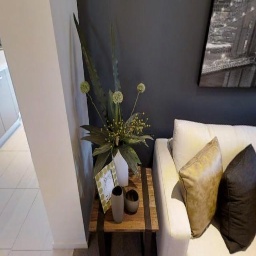}
        \end{minipage}
        \begin{minipage}{0.155\hsize}
            \centering
            \includegraphics[width=\linewidth]{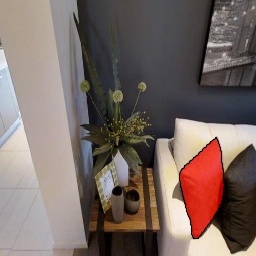}
        \end{minipage}
        \begin{minipage}{0.155\hsize}
            \centering
            \includegraphics[width=\linewidth]{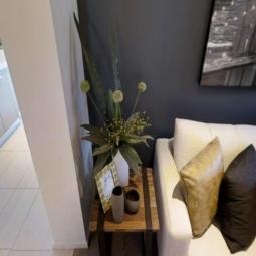}
        \end{minipage}
        \begin{minipage}{0.155\hsize}
            \centering
            \includegraphics[width=\linewidth]{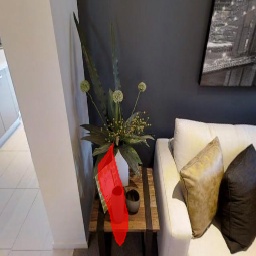}
        \end{minipage}
        \begin{minipage}{0.155\hsize}
            \centering
            \includegraphics[width=\linewidth]{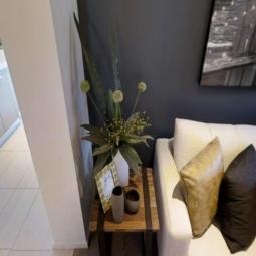}
        \end{minipage} 
        \begin{minipage}{0.155\hsize}
            \centering
            \includegraphics[width=\linewidth]{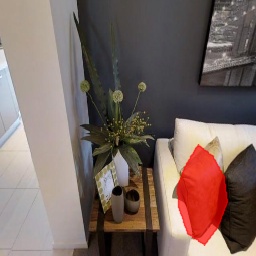}
        \end{minipage}
        (\small iii)
        \hspace{-2mm}
        \begin{minipage}{0.155\hsize}
            \centering
            \includegraphics[width=\linewidth]{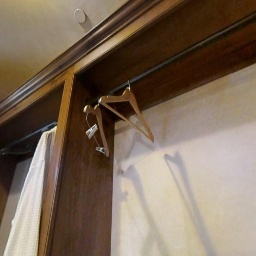}
        \end{minipage}
        \begin{minipage}{0.155\hsize}
            \centering
            \includegraphics[width=\linewidth]{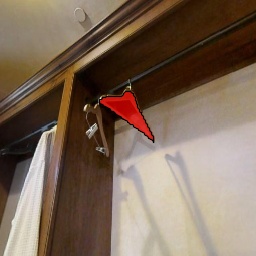}
        \end{minipage}
        \begin{minipage}{0.155\hsize}
            \centering
            \includegraphics[width=\linewidth]{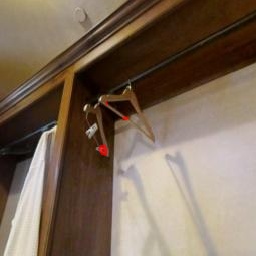}
        \end{minipage}
        \begin{minipage}{0.155\hsize}
            \centering
            \includegraphics[width=\linewidth]{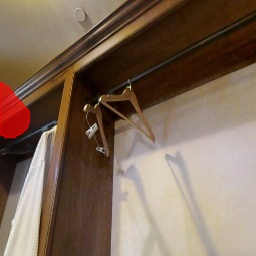}
        \end{minipage}
        \begin{minipage}{0.155\hsize}
            \centering
            \includegraphics[width=\linewidth]{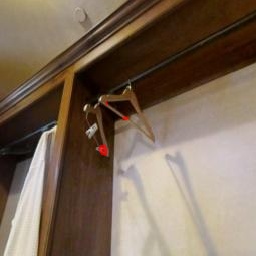}
        \end{minipage} 
        \begin{minipage}{0.155\hsize}
            \centering
            \includegraphics[width=\linewidth]{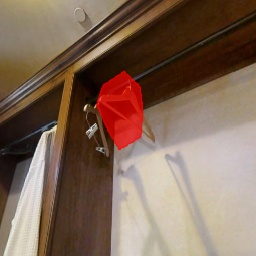}
        \end{minipage}\\
        \vspace{3mm}
        \hspace{5mm}
        \begin{minipage}{0.155\hsize}
            \centering
            (a) $\bm{x}_{\mathrm{img}}$
        \end{minipage}
        \begin{minipage}{0.155\hsize}
            \centering
            (b) GT
        \end{minipage}
        \begin{minipage}{0.155\hsize}
            \centering
            (c) LAVT\cite{yang2022lavt}
        \end{minipage}
        \begin{minipage}{0.155\hsize}
            \centering
            (d) SeqTR\cite{zhu2022seqtr}
        \end{minipage}
        \begin{minipage}{0.155\hsize}
            \centering
            (e) MDSM\cite{iioka2023mdsm}
        \end{minipage}
        \begin{minipage}{0.155\hsize}
            \centering
            (f) Ours
        \end{minipage}
    \vspace{-1mm}
    \caption{Qualitative results of successful and failure cases. (i) and (ii) show successful examples, and (iii) shows a failure example.
    The instructions for (i), (ii) and (iii) were as follows: ``In the 3rd level bathroom, there is a box of tissues to the left of the basin. Please fetch them''; ``Walk to the living room and fetch the leftmost pillow on the smaller white sofa, closest to the plant on the small table.'' and ``Go to the closet in the bedroom with the orange comforter and bring me the second hanger from the top.''}
    \label{fig:quantitysuc}
    \vspace{-1mm}
\end{figure*}
\begin{table}[t]
\centering
\caption{Comparison with baseline methods on the SHIMRIE-3D dataset. The bold numbers represent the highest values for each metric.}
{
\renewcommand{\arraystretch}{1}
{\normalsize 
\begin{tabular}{m{17mm}Wc{17mm}Wc{17mm}Wc{17mm}}
\hline
Model & mIoU [\%]$\uparrow$ & P@0.5 [\%]$\uparrow$ & P@0.7 [\%]$\uparrow$\\ \hline \hline
LAVT\cite{yang2022lavt}& 28.16 $\pm$ 2.85&  26.46 $\pm$ 4.01& 18.75 $\pm$ 3.29 \\
SeqTR \cite{zhu2022seqtr}& 21.84 $\pm$ 2.28&  17.87 $\pm$ 7.00& 5.16 $\pm$ 5.26 \\
MDSM\cite{iioka2023mdsm}& 24.36 $\pm$ 3.87&  22.49 $\pm$ 5.46& 13.71 $\pm$ 3.34 \\ \hline
Ours & \textbf{38.16 $\pm$ 2.46} &  \textbf{48.85 $\pm$ 2.70}& \textbf{22.29 $\pm$ 3.32} \\ \hline
\end{tabular}
\label{tab:quantitative}
}
}
\vspace{-5mm}
\end{table}
\begin{table*}[t]
\centering
\caption{Quantitative results on ablation studies. The bold numbers represent the highest values for each metric. The columns labeled VCI, OVA, SBAE, and PML indicate whether each module is included, as indicated by a check mark.}
{
{\normalsize 
\begin{tabular}{m{12mm}Wc{12mm}Wc{12mm}Wc{12mm}Wc{12mm}Wc{28mm}Wc{28mm}Wc{28mm}}
\hline
Model & VCI & OVA & SBAE & PML & mIoU [\%]$\uparrow$& P@0.5 [\%]$\uparrow$& P@0.7 [\%]$\uparrow$\\ \hline \hline
(i) & & \checkmark & \checkmark & \checkmark & 35.27 {$\pm$ 5.41}&  45.31 {$\pm$ 7.64}& 19.48 {$\pm$ 4.99}\\
(ii) & \checkmark & & \checkmark & \checkmark & 37.36 {$\pm$ 2.55}&  48.11 {$\pm$ 4.13}& \textbf{27.24 {$\pm$ 4.99}}\\
(iii) & \checkmark & \checkmark & & \checkmark & 31.77 {$\pm$ 0.92}&  37.86 {$\pm$ 2.06}& 14.00 {$\pm$ 4.28}\\
(iv) & \checkmark & \checkmark & \checkmark & & 33.07 {$\pm$ 3.44}&  41.04 {$\pm$ 6.74}& 20.42 {$\pm$ 8.18}\\
(v) & \checkmark & \checkmark & \checkmark & \checkmark &\textbf{38.16 {$\pm$ 2.46}} &  \textbf{48.85 {$\pm$ 2.70}}& 22.29 {$\pm$ 3.32} \\ \hline
\end{tabular}
\label{tab:abalation}
\vspace{-5mm}
}
}
\end{table*}
\section{
    Experimental Results
}
\vspace{-1mm}
\subsection{Quantitative Results}
Table~\ref{tab:quantitative} shows the quantitative results of the comparison of the baseline methods and proposed method. We conducted the experiments five times each. The averages and standard deviations of mIoU and P@$k$ ($k$=0.5, 0.7) are shown in the table.
Furthermore, the bold numbers in Table~\ref{tab:quantitative} represent the highest values for each metric.

As the baseline methods, we used MDSM\cite{iioka2023mdsm}, LAVT\cite{yang2022lavt}, and SeqTR\cite{zhu2022seqtr}.
We chose them as baseline methods for the following reasons.
We used MDSM because it has been successfully applied to the OSMI task.
Furthermore, we used LAVT and SeqTR because these have been successfully applied to the RES task, which is closely related to the OSMI-3D task.

We used mIoU and Precision at $k$ (P@$k$) as the evaluation metrics because they are standard metrics in RES tasks, closely associated with the  OSMI-3D task.
We chose mIoU as the primary metric. 
mIoU is defined as $\mathrm{mIoU} = ({1}/N_{s})\sum_{i=1}^{N_{s}}| Y_i \cap \hat{Y}_i |/| Y_i \cup \hat{Y}_i |$,
where $N_s$, $\hat{Y}_i$, and $Y_i$ denote the number of samples, and the sets of pixels corresponding to the predicted mask and GT mask, respectively, in the $i$-th sample.
P@$k$  is defined as $\mathrm{P}@k ={T_k}/N_{s}$,
where $T_k$ denotes the number of samples for which the IoU between a predicted mask and a GT mask exceeded the threshold $k$.

Table \ref{tab:quantitative} shows that the proposed method achieved an mIoU of 38.16\%, whereas LAVT, SeqTR, and MDSM were 28.16\%, 21.84\%, and 24.36\%, respectively.
The proposed method outperformed the best result for the baseline methods, which was obtained by LAVT, by 10.00 points. 
Table~\ref{tab:quantitative} also shows that P@0.5 for LAVT, SeqTR, MDSM, and the proposed method were 26.46\%, 17.87\%, 22.49\%, and 48.85\%, respectively.
From the above, the proposed method outperformed the highest performing LAVT in terms of P@0.5 by 22.39 points.
Similarly, the proposed method also outperformed the baseline methods in termes of P@0.7.
\subsection{Qualitative Results}
Fig.~\ref{fig:quantitysuc} shows the qualitative results.
In the figure, columns (a) and (b) represent $\bm{x}_{\mathrm{img}}$ and GT, respectively. 
Additionally, columns (c), (d), (e), and (f) represent the masks predicted by 
LAVT, SeqTR, 
MDSM and the proposed method, respectively.    
Fig.~\ref{fig:quantitysuc} (i)(ii) show successful examples.

Fig.~\ref{fig:quantitysuc} (i) shows an example in which the instruction was ``In the 3rd level bathroom, there is a box of tissues to the left of the basin. Please fetch them here.''
In this example, neither LAVT nor MDSM generated any masks, whereas SeqTR incorrectly masked the magazine. Conversely, the proposed method appropriately generated a mask for the tissue box, which demonstrates that it successfully identified the target object specified in the instruction.
In the example from Fig.~\ref{fig:quantitysuc} (ii), the instruction was ``Walk to the living room and fetch me the leftmost pillow on the smaller white sofa, the pillow closest to the plant on the small table.''
In this case, LAVT and MDSM also did not generate masks for the object, and SeqTR generated a mask for a different object on the table.
By contrast, the proposed method appropriately generated a mask for the beige cushion. We consider that the proposed method is capable of understanding referring expressions related to color and spatial relationships.

Fig.~\ref{fig:quantitysuc} (iii) illustrates a failure example.
Fig.~\ref{fig:quantitysuc} (iii) shows an example with the instruction sentence ``Go to the closet in the bedroom with the orange comforter and bring me the second hanger on top.''
In this example, both LAVT and MDSM generated an under-segmented mask for two hangers, and SeqTR masked unrelated areas.
Our method masked the hanger on the right-hand side, but the target object specified in the instruction sentence was the left hanger.
In this example, the phrase ``second hanger'' in the instruction sentence was ambiguous, which made it difficult to select a single target object.
\begin{figure}[t]
    \vspace{2mm}
    \captionsetup{}
    \centering
        \begin{minipage}{0.32\hsize}
            \centering
            \includegraphics[width=\linewidth]{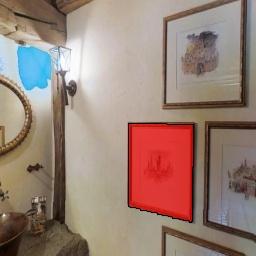}
        \end{minipage}
        \begin{minipage}{0.32\hsize}
            \centering
            \includegraphics[width=\linewidth]{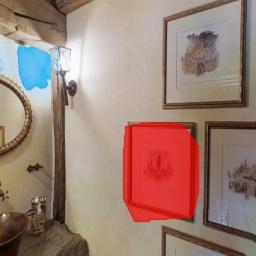}
        \end{minipage} 
        \begin{minipage}{0.32\hsize}
            \centering
            \includegraphics[width=\linewidth]{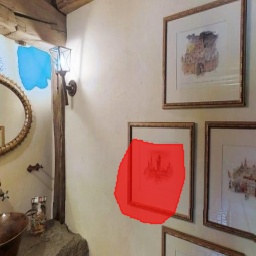}
        \end{minipage}\\
        \vspace{2mm}
        \begin{minipage}{0.32\hsize}
            \centering
            (a) GT
        \end{minipage}
        \begin{minipage}{0.32\hsize}
            \centering
            (b) Model (ii)
        \end{minipage}
        \begin{minipage}{0.32\hsize}
            \centering
            (c) Model (v)
        \end{minipage}
    \caption{The instruction sentence for this example was ``Go to the bathroom on level 1 and bring me the picture furthest to the left.'' In this case, the mask generated by Model (v) was slightly skewed toward the sink.}
    \label{fig:ablation}
    \vspace{-6mm}
\end{figure}
\vspace{-3mm}
\subsection{Ablation Studies}
We set the following four conditions for the ablation studies:

\textbf{VCI ablation}
We removed the VCI and assessed the contributions.
Table~\ref{tab:abalation} shows that the mIoU in for Model (i) was 35.27\%, which was 2.89 points lower than that for Model (v). 
P@0.5 and P@0.7 for Model (i) were also lower than that for Model (v).
From the above, VCI contributed to the improvement of performance.
This indicates that VCI enhanced the understanding of referring expressions.

\textbf{OVA ablation}
We investigated the performance of OVA by removing it.
Table~\ref{tab:abalation} indicates that the mIoU for Model (ii) was 37.36\%, which was 0.8 points lower than that for the Model (v). P@0.5 also decreased. 
By contrast, P@0.7 increased by 4.95 points.
This may indicate that information about objects that exist outside the camera's field of view was obtained by the OVA. 
However, using the OVA to obtain information around objects or the environment may inadvertently lead to focusing on nouns other than the target object. 
For example, Fig.~\ref{fig:ablation} shows an example where the instruction sentence was ``Go to the bathroom on level 1 and bring me the picture furthest to the left'' for which the mask generated by Model (v) was slightly skewed toward the sink. 
This is likely to have occurred because the model excessively focused on the word `bathroom' in the instruction, influenced by the feature of a sink related to `bathroom,' which was obtained through the OVA.
As a result, it is possible that the model could not focus strongly on the word `picture,' which was the target object.

\textbf{SBAE ablation}
To investigate the effectiveness of the SAM module in the SBAE, we removed it.
Table~\ref{tab:abalation} illustrates that Model (iii) achieved an mIoU of 31.77\%, which was 6.39 points lower than that of Model (v). It also scored lower in terms of P@0.5 and P@0.7.
This suggests that the SBAE enhanced the understanding of segment information about objects, thereby enabling the more appropriate prediction of object contours.

\textbf{PML ablation}
We investigated the implications for the performance of PML.
Table~\ref{tab:abalation} shows that the mIoU for Model (iv) was 33.07\%.
This result was 5.09 points lower than that for Model (v) and it was also lower in terms of P@0.5 and P@0.7.
This indicates that effective training was achieved using PML.
\vspace{-2mm}
\subsection{Error Analysis}
\vspace{-1mm}
In this study, we defined failed cases as those with an IoU lower than 0.5.
Based on this definition, the proposed method failed on 179 test samples.
We analyzed 100 samples with the lowest IoU values out of 179 samples of failure cases. Table~\ref{tab:error_analysis} describes the categories of the failure cases.

We roughly divided the cases into five types: 
\begin{itemize}
  \setlength{\parskip}{0.5mm} 
  \setlength{\itemsep}{0.2mm} 
  \item[(a)] Serious comprehension error\\
    This category includes failure cases in which our model incorrectly segmented a large part of objects that were not mentioned in the instruction. 
    For example, our model incorrectly segmented `wall' given the instruction ``Clean the decoration on the table.''
    This is presumably because our model failed to align the image and language.

  \item[(b)] Reference/exophora resolution error\\
    This category represents cases in which our model incorrectly segmented objects in the same category that were different from the target object. 
    For instance, our model improperly segmented ``the picture on the right-hand side'' following the instruction ``Bring the leftmost picture on the wall.'' 
    This is presumably because of  our model's failure to understand the referring expressions appropriately.

  \item[(c)] Segmentation of non-target objects\\
    This category refers to cases in which our model segmented non-target objects in the instructions. For example, `bed' was segmented given the instruction ``Fetch me a pillow on the bed.''

  \item[(d)] Hallucination in VCI\\
    The cases in this category involve multimodal LLM in VCI inappropriately describing the appearance and position of objects or non-existent objects.
    An example of this is, when there was no cushion in the room, the multimodal LLM generated the sentence ``The cushion on the left is white.''

  \item[(e)] Ambiguous instruction\\
    This category refers to cases in which the instructions included ambiguous expressions about the name or location of the target object, which made it difficult to identify the target object. 
    Suppose that the instruction ``Please bring the second hanger'' was provided and the image contained multiple hangers. It would remain unclear which hanger was being referred to.
\end{itemize}
\begin{table}[t]
    \vspace{2mm}
    \caption{Error analysis on failure cases.}
    \label{tab:error_analysis}
    {\normalsize
    \renewcommand{\arraystretch}{1}
    \begin{tabular}{m{5.5cm}|Wc{2cm}} \hline
    Errors & \# of Errors \\ \hline \hline
    Serious comprehension error & 43 \\ \hline
    Reference/exophora resolution error & 32  \\ \hline
    Segmentation of non-target objects & 13  \\ \hline
    \begin{tabular}[t]{@{}l@{}}Hallucination in VCI\end{tabular} & 10 \\ \hline 
    Ambiguous instruction   & 2 \\ \hline
    Total   & 100 \\ \hline
    \end{tabular}
    }
    \vspace{-6mm}
\end{table}
Table \ref{tab:error_analysis} indicates that the main bottlenecks were 
(a) and (b).
We consider that the reason for the former was that the model failed to ground referring expressions with their corresponding target objects. As a solution, we may be able to use SEEM \cite{zou2023segment}. 
SEEM performs open vocabulary panoptic segmentation, where textual features and visual prompt features are aligned in a joint visual-semantic space. 
To avoid focusing on irrelevant stuff or things, we can consider masking them out using this semantic labeling approach.
Furthermore, for the latter,
the model often misunderstood the spatial relationships between a target object and its surroundings. 
Additionally, there were some cases in which VCI failed to clearly describe the positional relationships between objects.
Therefore, a solution may be to improve the prompt so that it focuses on spatial relationships.
\vspace{-0.5mm}
\section{Conclusions}
\vspace{-1mm}

In this study, we focused on the OSMI-3D task, where models generated segmentation masks of the target object given an image of the indoor environment, 3D point clouds, and an instruction sentence related to object manipulation.
Our method outperformed the baseline methods on all standard metrics in the OSMI-3D task.
For future research, we plan to implement a semantic labeling approach to mask out irrelevant stuff, thereby ensuring that the focus remains on pertinent things.


\vspace{-1.0mm}
\section*{ACKNOWLEDGMENT}
\vspace{-1mm}
This work was partially supported by JSPS KAKENHI Grant Number 23H03478, JST Moonshot and NEDO.
\bibliographystyle{IEEEtran}
\bibliography{reference}

\end{document}